\documentclass[journal,10pt,draftclsnofoot,onecolumn]{IEEEtran}
\pdfoutput=1
\usepackage{url}
\usepackage{hyperref}
\usepackage{calc}
\usepackage[pdftex]{graphicx}
\usepackage[ left=0.75in,right=0.75in,top=1in,bottom=1in,%
                          footskip=.25in]{geometry}
\usepackage[table,x11names]{xcolor}
\usepackage{float}
\restylefloat{table}
\usepackage{array}
\usepackage{multirow}

\title{\bf Land Cover Classification from Multi-temporal, Multi-spectral Remotely Sensed Imagery using Patch-Based Recurrent Neural Networks}

\date{\vspace{-5ex}}

\author{
    \IEEEauthorblockN{Atharva Sharma\IEEEauthorrefmark{1}, Xiuwen Liu\IEEEauthorrefmark{1} and Xiaojun Yang\IEEEauthorrefmark{2} }
    \\\IEEEauthorblockA{\IEEEauthorrefmark{1}Department of Computer Science\\
    Florida State University, Tallahassee, Florida 32306-4530\\
    Email: as13an@my.fsu.edu and liux@cs.fsu.edu}\\
    \IEEEauthorblockA{\IEEEauthorrefmark{2}Department of Geography\\
    Florida State University, Tallahassee, Florida 32306-2190\\
    Email: xyang@fsu.edu}\\

}

\begin{document}

\maketitle
\begin{abstract} 
  Sustainability of the global environment is dependent on the
  accurate land cover information over large areas. Even with the
  increased number of satellite systems and sensors acquiring data
  with improved spectral, spatial, radiometric and temporal
  characteristics and the new data distribution policy, most existing
  land cover datasets were derived from a pixel-based single-date
  multi-spectral remotely sensed image with low accuracy. To improve
  the accuracy, the bottleneck is how to develop an accurate and
  effective image classification technique. By incorporating and
  utilizing the complete multi-spectral, multi-temporal and spatial
  information in remote sensing images and  considering their inherit
  spatial and sequential interdependence, we propose a new patch-based
  RNN (PB-RNN) system tailored for multi-temporal remote sensing
  data. The system is designed by incorporating distinctive
  characteristics in multi-temporal remote sensing data. In
  particular, it uses multi-temporal-spectral-spatial samples and
  deals with pixels contaminated by clouds/shadow present in the
  multi-temporal data series. Using a Florida Everglades ecosystem
  study site covering an area of 771 square kilometers, the proposed
  PB-RNN system has achieved a significant improvement in the
  classification accuracy over pixel-based RNN system, pixel-based
  single-imagery NN system, pixel-based multi-images NN system,
  patch-based single-imagery NN system and patch-based multi-images NN
  system. For example, the proposed system achieves 97.21\%
  classification accuracy while a pixel-based single-imagery NN system
  achieves 64.74\%.  By utilizing methods like the proposed PB-RNN
  one, we believe that much more accurate land cover datasets can be
  produced over large areas efficiently.
\end{abstract}

\section{ Introduction }
Land cover refers to the pattern of ecological resources and human activities dominating different areas of Earth's surface. It is a critical type of information supporting various environmental science and land management applications at global, regional, and local scales~\cite{meyer1994changes, Foley2005}. Given the importance of land cover information in global change and environmental sustainability research, there have been numerous efforts aiming to derive accurate land cover datasets at various scales (e.g,~\cite{Bartholome2005, Fry2011, Homer2007, Jin2013, Vogelmann2001, Gong2013}), mostly by using the remote sensing technology. However, even with the increased number of satellite systems and sensors acquiring data with improved spectral, spatial, radiometric and temporal characteristics and the new data distribution policy, most existing land cover datasets were derived from a pixel-based single-date multi-spectral remotely sensed imagery using conventional or advanced pattern recognition techniques such as random forests (RF)~\cite{ shi2016assessment}, neural networks (NNs)~\cite{Mas2008, Kavzoglu2003} and support vector machines (SVM)~\cite{yang2011parameterizing}. The real bottleneck is an accurate and effective image classification technique which can incorporate and utilize the complete multi-spectral, multi-temporal and spatial information available to provide land cover datasets for remote sensing images.\\

The remote sensing community has been well aware of the relevance of multi-temporal information in land cover mapping, but only limited exploitations have been attempted in actually utilizing such information, mostly from images with much coarse spatial resolutions such as MODIS (Moderate Resolution Imaging Spectroradiometer) time series (e.g., ~\cite{carrao2008contribution, vintrou2012crop, nitze2015temporal}). With the recently free availability of several major satellite remotely sensed datasets with much higher spatial resolutions acquired by the Landsat systems, ASTER (Advanced Spaceborne Thermal Emission and Reflection Radiometer), and Sentinels, exploiting multi-temporal information in land cover mapping is becoming more affordable and feasible. However, working with higher-resolution multi-temporal, multi-spectral imagery datasets is facing some crucial challenges mainly caused by the frequent occurrences of pixels contaminated by clouds or shadows ~\cite{Zhu2012} and the incompetency of some conventional pattern recognition methods~\cite{Gong2013}. Some pixel-based classification efforts attempted to use multiple cloud/shadow-free images acquired at different dates~\cite{bruzzone1999neural, bargiel2011multi}; but they failed to utilize the information of the inherit dependency of multi-temporal remotely sensed data and the invaluable spectral patterns associated with specific classes over time.\\

This work focuses on exploiting multi-temporal, multi-spectral and spatial information together for improving land cover mapping through the use of RNNs. Recently, RNNs  have been demonstrated to achieve significant results on sequential data and have been applied in different fields like natural language processing~\cite{mikolov2010recurrent, hinton2012deep, kalchbrenner2013recurrent}, computer vision~\cite{pinheiro2014recurrent, gregor2015draw, srivastava2015unsupervised}, multi-modal~\cite{karpathy2015deep, donahue2015long, graves2014neural} and robotics~\cite{mei2015listen}. RNNs have been applied on various applications such as language modeling, speech recognition, machine translation, question answering, object recognition, visual tracking, video analysis, image generation, image captioning, video captioning, self driving car, fraud detection, prediction models, sentimental classification, among others. Due to the inherit sequential nature of multi-temporal remote sensing data, such an effective technique could have significant impacts on multi-temporal remote sensing image classification. The remote sensing community has also attempted to utilize RNNs, but most of the existing efforts have been focused on the pixel-based change detection tasks~\cite{sauter2010spatio, qu2016deep, lyu2016learning}, with little on multi-temporal remote sensing image classification. Considering the inherit sequential interdependence of multi-temporal remote sensing data and the spatial relation of a pixel to its neighbourhood, we have developed a patch-based RNN (PB-RNN) system for land cover classification from a Landsat-8 OLI (Operational Land Imager) time series that are freely available from USGS~\cite{Data_Access2016}. We targeted medium-resolution Landsat imagery because of their overwhelming use as the primary data for land cover classification and environmental sustainability research.
%The data acquired by Landsat programs provide the longest continuous observations of Earth's surface from space. In 
%particular, the Landsat system offers a rich archive of highly calibrated, multi-spectral data of global coverage that recently 
%becomes available at no charge from the USGS EROS Data Center. Landsat8 is the latest edition of the Landsat programs 
%which images the entire earth every 16 days; therefore, there is an addition of a datum point to the Landsat 8 sequential data %belonging to all the desired locations after every 16 days~\cite{Data_Access2016}.
Our proposed method also includes a component to deal with pixels contaminated by clouds/shadow present in the multi-temporal data series. Using a test site in a complicated tropical area in Florida, our proposed PB-RNN system has achieved a significant improvement in the classification accuracy over pixel-based RNN system, pixel-based single-imagery NN system, pixel-based multi-images NN system, patch-based single-imagery NN system and patch-based multi-images NN system. \\   

The remainder of this paper is organized as follows. In Section \ref{sec2}, we describe RNNs customized for remote sensing applications. In Section \ref{sec3}, we describe our proposed PB-RNN system to map land cover types from multi-temporal, multi-spectral  remotely sensed images. In Section \ref{sec4}, we present our experimental results and also compare them with the outcomes from pixel-based RNN system, pixel-based single-imagery NN system, pixel-based multi-images NN system, patch-based single-imagery NN system and patch-based multi-images NN system. Finally, Section \ref{sec5} summarizes the major findings and discusses some issues for further research.

\section{Recurrent Neural Networks (RNNs)}\label{sec2}
In conventional multilayer NNs, all the inputs belonging to a particular sequence or time series are considered independent to each other and are associated with different parameters present in the network. Due to the above properties, the standard multilayer NNs are limited when dealing with sequential data because these networks are impotent to utilize the inherit dependency between the sequential inputs. On the other hand, RNNs can deal with sequential data efficiently~\cite{Goodfellowetal}. RNNs consider the dependency between the sequential inputs; and use the same function and same set of parameters at every time step. Using RNN, a sequence of vector x can be processed by applying a recurrence formula at every time step.
\begin{equation}\label{eq:1}
h_{t} = f( h_{t-1},x_{t};\Theta )
\end{equation}
In Equation \ref{eq:1}, $h_{t}$ represents the new state which can be obtain using some non-linear function f with old state $h_{t-1}$, input vector $x_{t}$ and set of parameters  $\Theta$ as the inputs.
In case of simple RNN, the recurrent equations are as follows:
\begin{equation}\label{eq:2}
h_{t} = f_{h}( W_{x}x_{t} +W_{h}h_{t-1} + b_{h} )
\end{equation}
\begin{equation}\label{eq:3}
y_{t} = f_{y}( W_{y}h_{t} + b_{y} )
\end{equation}
where input, hidden state and output vector of RNN at time t are represented by $x_{t}$, $h_{t}$ and $y_{t}$, respectively, and $W_{x}$, $W_{h}$, $W_{y}$, and $b_{h}$, $b_{y}$ are learnable parameters.\\
An RNN architecture can be designed in various ways based on their input/output. One-to-sequence, in which a single input is used to generate a sequence as an output for example image captioning~\cite{karpathy2015deep}. Sequence-to-sequence, where a sequence of data is used to generate a sequence as an output (e.g., machine translation and video classification on frame level)~\cite{kalchbrenner2013recurrent, srivastava2015unsupervised}. Sequence-to-one architecture, which takes sequential data as an input to produce a single output (e.g., sentimental classification, automatic movie review)~\cite{tang2015document}. Land cover classification using our proposed approach to remote sensing imagery bears similar property to the sequence-to-one architecture, where multi-temporal data sequence can be used as an input to classify the desired location into one of the defined classes. Each sample defines a sequence of patches of size $X \times Y \times Z$ labeled using the center pixel, over the time interval of the sequence length; where each datum point in the sequence represents a flatten p-dimensional vector of length p, whose size is $X \times Y \times Z$  at a specific time. However, with the small sequence length, the standard RNN version with non-linear function like sigmoid or hyperbolic tangent works pretty well but the standard version is not capable of dealing with long-term dependencies because of the gradient vanishing and exploding problem while backpropagating in the training stage of RNN. Fortunately, a special kind of RNN named,  long short term memory networks (LSTMs) can overcome the gradient vanishing and exploding problem and are capable of dealing with long-term dependencies. The proposed system has also used LSTM to avoid gradient vanishing and exploding problem.\\
\begin{figure}
\centering
\includegraphics[width=6.0in]{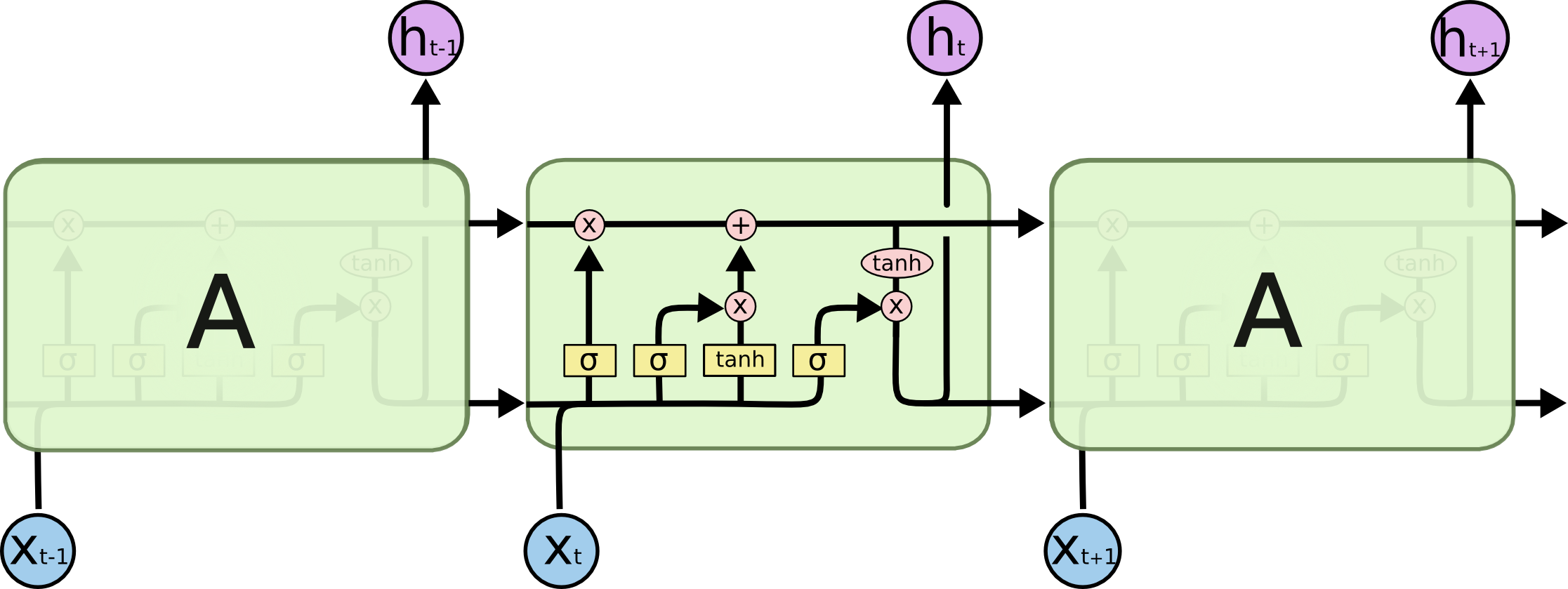}
 \caption[LSTM Architecture]{ LSTM Architecture\footnotemark }\label{lstm}
\end{figure}
\footnotetext{Obtained from http://colah.github.io/posts/2015-08-Understanding-LSTMs/ }
Basically, LSTMs (see Figure~\ref{lstm})  work using  a gating mechanism with a memory cell. Hidden state is represented as a vector and is calculated using three gates named as input $i$, forget $f$ and output $o$ gates. All of these gates use sigmoid function, which restrict the value of these vectors between 0 and 1. By element-wise multiplying these gates with another vector, these gates define the proportion of other vector they allow to let through. The input gate defines the proportion of newly computed state for the current input it allows to let through. The forget gate controls how much of the previous state it allows to be considered. The output gate decides the proportion of the internal state it passes to the external network. Memory cell $c$ defines the combination of previous memory and the new input. Memory cell $c$ at time $t$ is calculated by combining the element-wise multiplication of the memory cell at previous time point with the forget gate, and the element-wise multiplication of newly computed state with input gate. Finally, LSTM calculates hidden state $h_{t}$ at time $t$ by multiplying the memory cell with the output gate element-wise. The whole architecture can be defined using following equations:
\begin{equation}
i_{t} =sigmoid( W_{x1}x_{t} + W_{h1}h_{t-1} )
\end{equation}
\begin{equation}
f_{t} = sigmoid( W_{x2}x_{t} + W_{h2}h_{t-1} )
\end{equation}
\begin{equation}
o_{t} = sigmoid( W_{x3}x_{t} + W_{h3}h_{t-1} )
\end{equation}
\begin{equation}
g_{t} = tanh( W_{x4}x_{t} + W_{h4}h_{t-1} )
\end{equation}
\begin{equation}
c_{t} = f_{t} \odot c_{t-1} + i_{t} \odot g_{t}
\end{equation}
\begin{equation}
h_{t} =o_{t} \odot  tanh( c_{t} )
\end{equation}
where input, forget and output gates at time $t$ are represented by $i_{t}$, $f_{t}$ and $o_{t}$, respectively; $c_{t}$ represents the cell $c$ at time $t$. Both, current input vector $x_{t}$ at time $t$ and previous hidden state $h_{t-1}$ are used as the inputs in the LSTMs. $W_{x1}$, $W_{x2}$, $W_{x3}$, $W_{x4}$, and  $W_{h1}$, $W_{h2}$, $W_{h3}$, $W_{h4}$ are learnable parameters;  The element-wise multiplication is denoted by $\odot$ symbol.\\

These gates with memory cell allow LSTMs to analyze the long dependencies by going deep in time without facing gradient vanishing and exploding problem. Using the input gate, LSTMS are also capable of avoiding cloud/shadow points present in the sequence.

\section{ Proposed Patch-Based RNN System (PB-RNN) for Land Cover Classification}\label{sec3}
The proposed system is adapted for the land cover classification using complete multi-temporal, multi-spectral and spatial information together for remotely sensed imagery. While the proposed method is generic and should work for all the multi-temporal-spectral remote sensing imagery, we have tested the new method on Landsat images. Below, we define the features used and the architecture adopted.

\subsection{ Multi-Temporal-Spectral Data }
Multi-temporal-spectral data are generated in two phases; firstly, we extracted multi-spectral layer stacks out of Landsat images and then a series of layer stacks are combined together to get the final product. In order to convert a Landsat 8 imagery into a multi-spectral layer stack, we have calculated the top-of-atmosphere (TOA) reflectance values associated with the pixels from the scaled digital numbers (DN) belonging to all the OLI bands (except the panchromatic band). TOA reflectance values can be obtained by rescaling and correcting the default 16-bit unsigned integer format DN values using radiometric (reflectance) rescaling coefficients and Sun angle provided in the MTL file present with Landsat 8 product~\cite{Product2016}. In the multi-spectral layer stack, each pixel is represented as a vector of length $Z$ consisting TOA reflectance values belonging to $Z$ OLI bands. Landsat 8 program images the entire earth every 16 days; so, if we consider a time series of $N$ images belonging to the desired location then the time interval between any two consecutive images in the series is equal to 16 days. Before generating multi-temporal-spectral data, all the $N$ images in the series are converted into multi-spectral layer stacks explained above, individually. Finally, we club these $N$ multi-spectral layer stacks belonging to a series of $N$ images together. 

\subsection{ Multi-Temporal-Spectral-Spatial Samples }
In the above mentioned multi-temporal-spectral data, both temporal and spectral information is automatically fused together. However, our proposed system is trying to use the complete multi-spectral, multi-temporal and spatial information available for land cover classification. Therefore, in order to include the spatial information in our samples, we have to extract each sample as a sequence of patches from the above mentioned multi-temporal-spectral data instead of a sequence of $N$ vectors of length $Z$ representing TOA reflectance associated to $Z$ OLI bands belonging to a single pixel. So, each sample is extracted as a sequence of $N$ patches of size $X \times Y \times Z$ labeled using the center pixel where $N$ specifies the sequence length, $X$ the width, $Y$ the height and $Z$ the number of bands, respectively. In this representation, values of the sequence length $N$ and consisting patches of size $X \times Y \times Z$ vary according to the problem and imagery type but the structure of the multi-temporal-spectral-spatial sample remains the same. For the implementation purpose, patches are flattened into vectors of length $X \times Y \times Z$. In the proposed architecture, a series of 23 images are considered and patches of size $3 \times 3 \times 8$ representing TOA reflectance values of 8 OLI bands (except the panchromatic band) of $3 \times 3$ window labeled using the center pixel are flattened into vectors of length 72 (3*3*8). Each sample defines a sequence of 72-dimensional TOA reflectance vectors which consist of spectral and spatial information belonging to the center pixel location of a distinct $3 \times 3$ window over the whole year.  Each datum point in the sequence represents a TOA reflectance vector at a specific time and there is a time interval of 16 days between any two consecutive points in the sequence; in order to cover the whole year, there are 23 datum points present in the sequence. 

\subsection{ Cloud/Shadow Datum Points }
In order to deal with cloud/shadow datum points in the sequence, cloud/shadow masks are generated for all the 23 Landsat images using the Fmask Algorithm~\cite{Zhu2012}, individually. These masks are then used to locate all the cloud/shadow points present in the multi-spectral layer stacks corresponding to the Landsat images. TOA reflectance vectors of the cloud/shadow locations are set to zero vector. Input gate in the LSTM cell helps to deal cloud/shadow datum vectors by restricting these zero vectors to let through; so that, these cloud/shadow datum points won't effect the information from the clear datum points. 

\subsection{ Training Samples }
In order to be consistent with all the other pixel-based and patch-based neural networks used in this paper, image acquired on March 30, 2014 with less than two percent cloud/shadow cover on our test site is used to extract training samples location. In addition, we impose two constraints. First, there should be no cloud/shadow pixel present in the patch at this date. Second, all the boundary patches should be avoided. Eighty percent samples which satisfy the above constraints are extracted separately from all the distinct classes.
\subsection{ Architecture }
\begin{figure}
\centering
\includegraphics[width=6.0in]{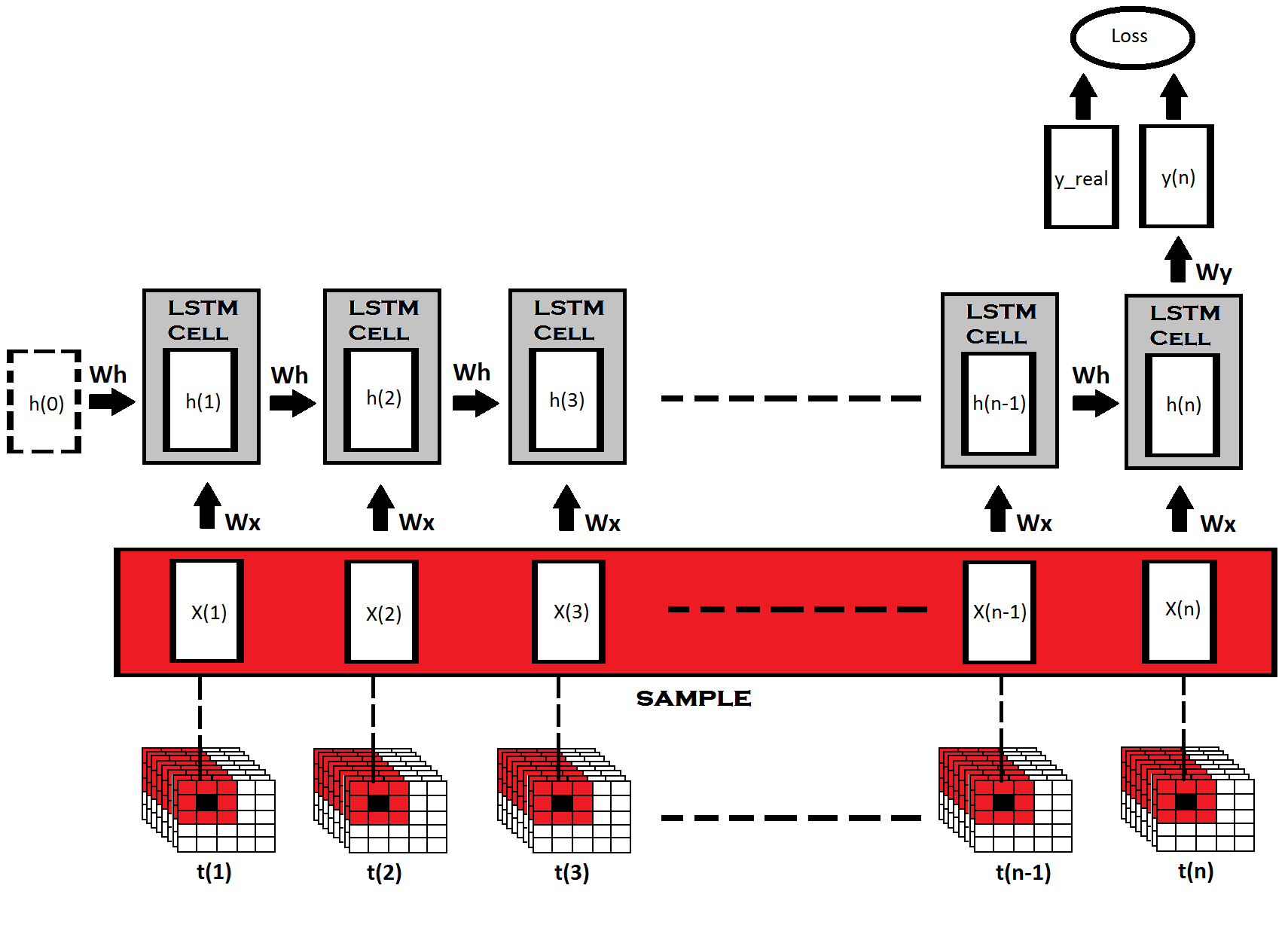}
 \caption{ Architecture of the proposed patch-based RNN (PB-RNN) system for land cover classification }\label{lstm_arch}
\end{figure}

Figure~\ref{lstm_arch} illustrates the general overall architecture of the proposed PB-RNN system for land cover classification. Sample X is shown in red; represents a sequence of n vectors ( X(1) to X(n) ) of equal length obtained from time point t(1) to point t(n), respectively. In our experiment, each vector of length 72 (3*3*8) in the sequence is extracted by flattening a patch of size $3 \times 3 \times 8$  representing TOA reflectance values associated to 8 OLI bands of $3 \times 3$ window labeled using the center pixel at a particular time point in the time series and the length of the sequence is 23 to cover the whole year. The softmax layer generates a probability distribution over the eight classes, using the output from the LSTM cell at the 23rd time step as its input.\\ 
%In our experiment, each vector in the sequence is of length 72 representing TOA reflectance values associated to 8 OLI bands at %particular time point in the time series and the length of the sequence is 23 to cover the whole year. The softmax layer generates

The proposed system has sequence-to-one architecture, where flattened patch vectors sequence is used as an input to classify the desired location into one of the defined classes. In order to implement this network, we have used Google's tensorflow (an open source software library for machine intelligence)~\cite{tensorflow2015-whitepaper} and Quadro K5200 GPU. The proposed network minimizes the cross entropy using the ADAM optimizer~\cite{kingma2014adam} with a learning rate of 0.0001. The ADAM optimizer is a first-order gradient-based optimization algorithm of stochastic objective functions, stochastic gradient descent proves to be a very efficient and effective optimization method in recent deep learning networks (e.g. ~\cite{Krizhevsky2012, deng2013recent, hinton2006reducing, hinton2012deep}).\\

Using LSTM cell recurrent network, a sequence of vectors X(1:n) can be processed by applying a recurrence formula at every time step. In order to calculate the current hidden state h(t) at time $t$, LSTM cell takes current input vector X(t) from the input sequence and previous hidden state h(t-1) as inputs. Initial hidden state h(0) is initiated as a zero state. Current state depends on all the relevant previous states and input vectors in the sequence; the irrelevant information is controlled by the gate mechanism by resisting it to let through. The proposed system has a fixed length of 23 for the input sequence; so, h(23) is the final hidden state here, and $W_{x}$ and $W_{h}$ are the learnable weight matrices, which remain the same at every time step.

\section { Experimental Results and Comparisons} \label{sec4}
\subsection{Test Site}

We chose to implement the proposed method on a test site within the Florida Everglades ecosystem;
this ecosystem has attracted international attention for the
ecological uniqueness and fragility. It is comprised of a wide variety
of sub-ecosystems such as freshwater marshes, tropical
hardwood hammocks, cypress swamps, and mangrove swamps~\cite{davis1994everglades}. Such diverse ecological
types make the Everglades an ideal site to test the reliability and
robustness of this new system. A series of 23 Landsat8 images was used in our study, where the time interval between any two consecutive images in the series is 16 days. The first image of the series was acquired on February 10, 2014 and the last  image on January 28, 2015 (Path 16; Row 42); we extract a subset with a
size equivalent to 771 square kilometers ($864\times 991$ pixels) from the whole image as our test site. In order to do single imagery classification, we have used the image acquired on March 30, 2014 with less than two percent cloud/shadow cover on our test site. For multi-images based classification, we have used 4 images acquired on February 10, 2014; March 14, 2014; March 30, 2014; and January 28, 2015 with less than four percent cloud/shadow cover for all of them on our test site.

In order to create a reference map for our research area, we obtained ancillary data from the Florida Cooperative Land
Cover Map first and performed correction by comparing  it with GPS-guided field observations and the high-resolution images from Google Earth. We have used this reference map to generate training samples and
perform  accuracy assessments~\cite{lo1998influence}. Using the ancillary data,
we adopt a mixed Anderson Level 1/2 land-use/land-cover classification
scheme~\cite{anderson1976land} with eight classes (see
Table~\ref{scheme}). Based on our training sample extraction constraints, we are able to generate training samples of size $23\times 72$ for the eight classes, where 23 is the sequence length and 72 (3*3*8) is the flattened patch vector size; the details are shown in Table~\ref{scheme}. 

\begin{table*}\caption{\bf Land cover classification scheme and training sample size.}\label{scheme}
\begin{center}
 \scalebox{0.65}{
    \begin{tabular}{ | l | l |p{12.5cm}|l|}
    \hline
    {\bf No.} & {\bf Class Name} & {\bf Description} & {\bf Training} \\ 
 {} & {} & {} & {\bf  Samples} \\ \hline
    1 & High Intensity Urban & Commercial, industrial, institutional constructions with large roofs. Large open spaces and large transportation facilities. Residential areas with impervious surfaces more than half of the total cover. & 115879\\ \hline
    2 & Low Intensity  Urban & Residential areas with impervious surfaces less than half of the total cover. Smaller urban service buildings, such as detached stores and restaurants, and state highways. & 63726\\ \hline
    3 & Barren Land & Urban areas with low percentages of constructed materials, vegetation, and low level of impervious surfaces, including bare soil lands, beaches. & 14090  \\ \hline
    4 & Forest & Herbaceous cover, trees, trees remain green throughout the year, some wetland evergreen forests included. & 90255  \\  \hline 
    5 & Cropland & Crops and pastures with vegetation coverage mixed with bushes, small amount fallow land. & 77150 \\ \hline
    6 & Woody Wetland & Cypress/tupelo, strand swamp, other coniferous wetland, mixed wetland hardwoods, mangrove swamp. & 147851\\ \hline
    7 & Emergent Herbaceous Wetland & Freshwater non-forested wetland, prairies/bogs, freshwater marshes, wet prairies, saltwater marsh.  & 39103\\ \hline
    8 & Water & Streams, canals, lakes, ponds, bays. & 116951\\ \hline
    \end{tabular}
}
\end{center}
\end{table*}

\subsection{Experimental Results/Comparisons}

In this section, we present the experimental results of the proposed PB-RNN system and compare them from those from pixel-based RNN system, pixel-based single-imagery NN system, pixel-based multi-images NN system, patch-based single-imagery NN system and patch-based multi-images NN system. By comparing  pixels directly in  each of the classification maps from the six different networks for the whole area to the reference map, pixel-based single-imagery NN system achieves accuracy of 62.82\%, pixel-based multi-images NN system 63.57\%, patch-based single-imagery NN system 73.07\%, patch-based multi-images NN system 73.95\%, pixel-based RNN system 84.09\% and PB-RNN system 96.92\%. However, in order to perform quantitative evaluation of the classification results generated by six different neural networks to determine overall and individual category classification accuracy, we have done accuracy assessment using the method described by Congalton~\cite{congalton1991review}. Specifically for each method, an error matrix is generated using the weighted random stratified sampling and then the overall accuracy (OA), Overall kappa (KAPPA), Producer's accuracy (PA), User's accuracy (UA) and conditional kappa are calculated based on the error matrix~\cite{Jensen2015}.\\

\begin{figure*}
\centering
 \includegraphics[scale=0.375]{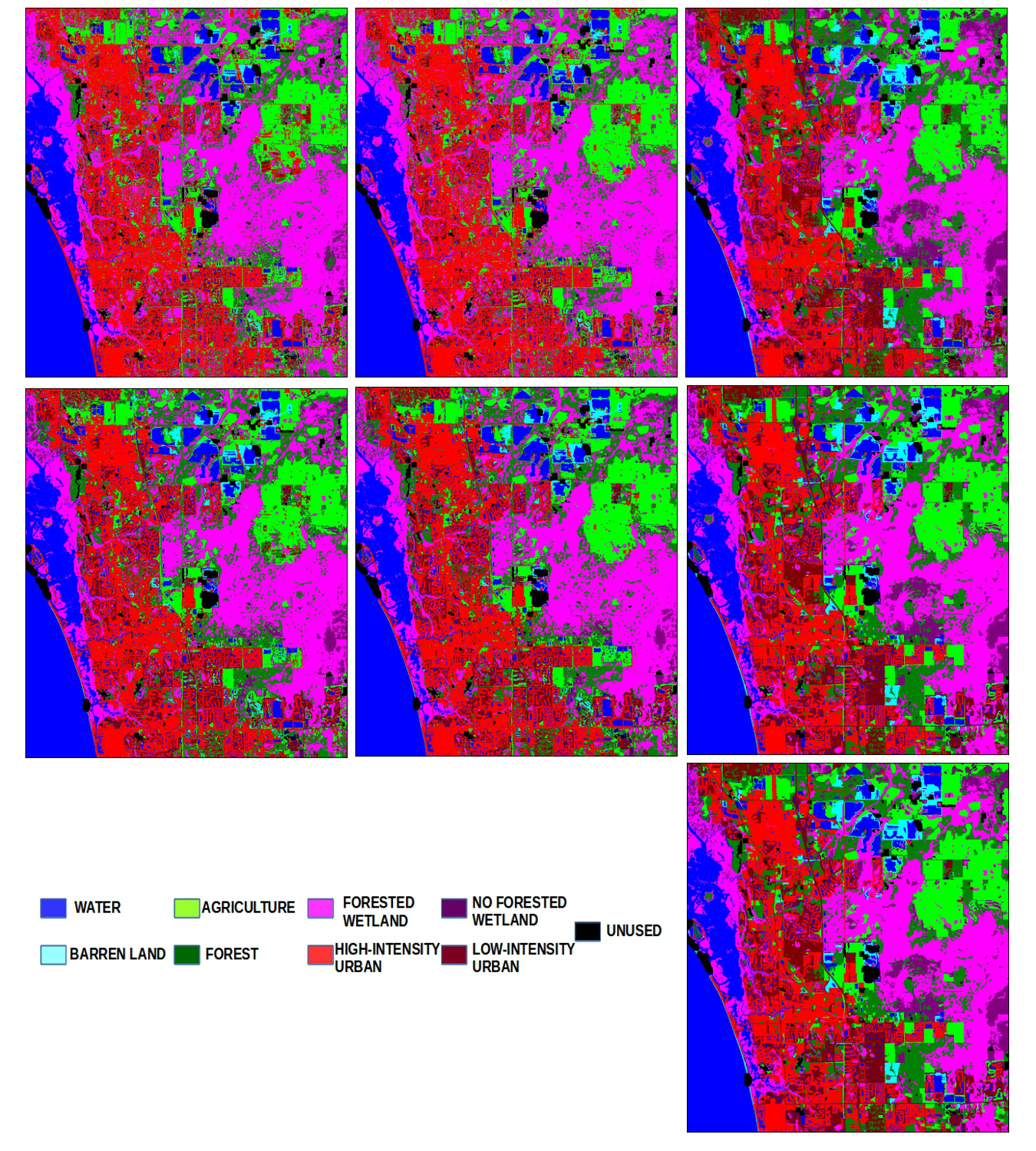}
 \caption{\bf Classification results for the study site from six different methods: (a) Upper Left: Pixel-based single-imagery NN system, (b) Upper Center:  Pixel-based multi-images NN system, (c) Upper Right: Pixel-based RNN system, (d) Middle Left: Patch-based single-imagery NN system, (e) Middle Center: Patch-based multi-images NN system, (f) Middle Right: Proposed patch-based RNN (PB-RNN) System and (g) Lower Right: Reference map.  }\label{refer_comp}
\end{figure*}

\begin{table*}\caption{\bf Error Matrix using the Patch-Based RNN (PB-RNN) System.}\label{matrix_rnn}
\begin{center}
   \scalebox{0.545}{
    \begin{tabular}{  | c |  c | c |  c | c |  c |  c | c |  c | c | c | c | c |}
    \hline
     \multirow{4}{*}{Classified Data}& \multicolumn{9}{|c|}{Reference Data } &  \multicolumn{3}{|c|}{Accuracy and Conditional Kappa}\\  \cline{2-13}
      & { High} & { Low}  & { Barren} & { Forest} & {Crop} &{ Woody}& { Emergent }  &{ Water}& { Row}  & { Producer's} & { User's} & { Conditional} \\ 
     & { Intensity} & { Intensity} & { Land} & {} & {land} &{  Wetland}& {Herbaceous}   &{ } & { Total} & { Accuracy} & {Accuracy} & { Kappa} \\  
     & { Urban} & { Urban}  &{ } & {} &{} & {} & { Wetland} & {} & { } & { (PA \%)} & {(UA \%)} & { } \\  \hline
      High Intensity Urban & {\bf 154} & 2  & 0 & 0  & 1 & 0 & 0 & 1 &158 & 97.47 & 97.47 & 0.97 \\ \hline
     Low Intensity Urban  & 3 & {\bf 82} & 0 & 0 & 0 & 1 & 0   & 1 & 87 & 96.47 & 94.25 & 0.94\\ \hline
     Barren Land & 0 & 0  & {\bf 50} & 0 & 0 & 0 & 0  & 0 & 50  & 98.04 & 100 & 1.00\\ \hline
    Forest  & 0 & 0 & 0 & {\bf 118} & 1 & 1 & 0  & 0 & 120  & 95.93 & 98.33 & 0.98\\ \hline
   Cropland & 0 & 0 & 0 & 1 & {\bf 103} & 0 & 1 & 0 & 105 & 98.1 & 98.1 & 0.98\\ \hline
   Woody Wetland  & 0 & 0 & 0 & 3 & 0 & {\bf 195} & 2  & 2 & 202 & 97.99 & 96.53 & 0.96\\ \hline
    Emergent Herbaceous   & \multirow{2}{*}0 & \multirow{2}{*}0 &  \multirow{2}{*}1 & \multirow{2}{*}0 & \multirow{2}{*}0 & \multirow{2}{*}2 & \multirow{2}{*}{\bf 48}   & \multirow{2}{*}0 & \multirow{2}{*}{51}  & \multirow{2}{*}{94.12} & \multirow{2}{*}{94.12} & \multirow{2}{*}{0.94}\\ 
      Wetland &  &  &  &  &  &  &  &  &   &  &  & \\ \hline
       Water  & 1 & 1 & 0 & 1 & 0 & 0 & 0 & {\bf 155} & 158 & 97.48 & 98.1 & 0.98 \\ \hline

    Column Total  & 158 & 85 & 51 & 123 & 105 & 199 & 51  & 159 & {\bf 931} & \multicolumn{3}{|c|}{}\\ \hline
 \multicolumn{13}{|c|}{ Overall Accuracy(OA): {\bf 97.21\%};  Overall Kappa(KAPPA): {\bf 0.967} } \\  
    \hline
     
    \end{tabular}
}
\end{center}
\end{table*}

\begin{table*}\caption{\bf Summary of the accuracy assessment for the classification results produced by the new patch-based RNN system (PB-RNN), pixel-based RNN system (Pix RNN), pixel-based single-imagery NN system (Pix Single), pixel-based multi-images NN system (Pix Multi), patch-based single-imagery NN system (Patch Single) and patch-based multi-images NN system (Patch Multi).
}\label{comparison}
\begin{center}
   \scalebox{0.8}{
    \begin{tabular}{ | c |  c | c |  c| c | c |  c| c | c | }
\hline
   \multirow{2}{*}{Land Cover Class} & \multicolumn{8}{|c|}{ Conditional Kappa } \\ \cline{2-9}
                                                                      & Pix Single & Pix Multi & Patch Single & Patch Multi  & Pix RNN & PB-RNN  &  Mean  &  Standard Deviation\\ \hline
        High Intensity Urban & 0.54 & 0.57 &  0.69 & 0.71 & 0.81 & 0.97 &  0.72 & 0.16  \\ \hline
      Low Intensity Urban & 0.54 & 0.54  &  0.66 &  0.67&  0.78 &  0.94 & 0.69 &  0.15  \\  \hline 
     Barren Land & 0.52 &  0.66 &  0.60 &   0.64 &  0.85 &  1.00 &    0.71 &  0.18  \\ \hline
    Forest & 0.48 &  0.53 &  0.52 &  0.56 &  0.79 &  0.98 &  0.64 &  0.20  \\ \hline
    Cropland & 0.57 &  0.57 &  0.78 &  0.78 &  0.95 &  0.98 &  0.77 &  0.18  \\ \hline
      Woody Wetland &  0.53 &  0.51 &  0.69 &  0.72 &  0.89 &  0.96 &  0.72 &  0.18  \\ \hline
    Emergent Herbaceous Wetland & 0.57 &  0.63 &  0.72 &  0.83 &  0.75 &  0.94 &  0.74 &  0.13 \\ \hline
      Water & 0.86 &  0.88 &  0.94 &  0.95 &  0.94 &  0.98 &  0.93 &  0.05  \\ \hline
      Mean-Kappa & 0.58 &  0.61 &  0.70 &   0.73 &  0.84 &  0.97 & \multicolumn{2}{|c|}{\multirow{4}{*}{}} \\ \cline{1-7} 
     Standard Deviation & 0.12 &  0.12 &  0.12 &  0.12 &  0.08 &  0.02 &  \multicolumn{2}{|c|}{} \\ \cline{1-7} 
     {\bf Overall Accuracy(\%)} & 64.74 & 66.40 & 75.54 & 77.63 & 87.65 & 97.21 &  \multicolumn{2}{|c|}{} \\ \cline{1-7} 
     {\bf Overall Kappa} & 0.58 & 0.60 & 0.71 & 0.74 & 0.86 & 0.97 &  \multicolumn{2}{|c|}{} \\  \hline
   
    \end{tabular}
}
\end{center}
\end{table*}

\begin{table*}\caption{\bf Error Matrix using the Pixel-Based RNN System.}\label{matrix_rnn_pixel}
\begin{center}
   \scalebox{0.545}{
    \begin{tabular}{  | c |  c | c |  c | c |  c |  c | c |  c | c | c | c | c |}
    \hline
     \multirow{4}{*}{Classified Data}& \multicolumn{9}{|c|}{Reference Data } &  \multicolumn{3}{|c|}{Accuracy and Conditional Kappa}\\  \cline{2-13}
      & { High} & { Low}  & { Barren} & { Forest} & {Crop} &{ Woody}& { Emergent }  &{ Water}& { Row}  & { Producer's} & { User's} & { Conditional} \\ 
     & { Intensity} & { Intensity} & { Land} & {} & {land} &{  Wetland}& {Herbaceous}   &{ } & { Total} & { Accuracy} & {Accuracy} & { Kappa} \\  
     & { Urban} & { Urban}  &{ } & {} &{} & {} & { Wetland} & {} & { } & { (PA \%)} & {(UA \%)} & { } \\  \hline
      High Intensity Urban & {\bf 137} & 13  & 0 & 4  & 0 & 1 & 0 & 8 &163 & 91.33 & 84.05 & 0.81 \\ \hline
     Low Intensity Urban  & 4 & {\bf 69} & 1 & 6 & 1 & 2 & 0   & 3 & 86 & 77.53 & 80.23 & 0.78\\ \hline
     Barren Land & 2 & 1  & {\bf 43} & 1 & 2 & 0 & 0  & 1 & 50  & 86 & 86 & 0.85\\ \hline
    Forest  & 1 & 1 & 1 & {\bf 101} & 5 & 8 & 3  & 3 & 123  & 80.16 & 82.11 & 0.79\\ \hline
   Cropland & 1 & 0 & 0 & 1 & {\bf 92} & 1 & 0 & 1 & 96 & 90.2 & 95.83 & 0.95\\ \hline
   Woody Wetland  & 2 & 3 & 1 & 10 & 0 & {\bf 186} & 1  & 1 & 204 & 90.73 & 91.18 & 0.89\\ \hline
    Emergent Herbaceous   & \multirow{2}{*}1 & \multirow{2}{*}1 &  \multirow{2}{*}2 & \multirow{2}{*}3 & \multirow{2}{*}1 & \multirow{2}{*}6 & \multirow{2}{*}{\bf 45}   & \multirow{2}{*}0 & \multirow{2}{*}{59}  & \multirow{2}{*}{91.84} & \multirow{2}{*}{76.27} & \multirow{2}{*}{0.75}\\ 
      Wetland &  &  &  &  &  &  &  &  &   &  &  & \\ \hline
       Water  & 2 & 1 & 2 & 0 & 1 & 1 & 0 & {\bf 143} & 150 & 89.38 & 95.33 & 0.94 \\ \hline

    Column Total  & 150 & 89 & 50 & 126 & 102 & 205 & 49  & 160 & {\bf 931} & \multicolumn{3}{|c|}{}\\ \hline
 \multicolumn{13}{|c|}{ Overall Accuracy(OA): {\bf 87.65\%};  Overall Kappa(KAPPA): {\bf 0.855} } \\  
    \hline
     
    \end{tabular}
}
\end{center}
\end{table*}

The proposed PB-RNN system uses samples, which are in the form of time series of 72-dimensional (3*3*8) flattened TOA reflectance patch vectors of size $23 \times 72$. In case of pixel-based RNN system instead of a patch of size $3 \times 3 \times 8$, each vector in the sequence is representing only the center pixel vector of each patch and this pixel vector is of length 8, containing the TOA reflectance values of 8 OLI bands at that location; so, samples are of size $23 \times 8$. Patch-based single-imagery NN system uses 72-dimensional (3*3*8) flattened TOA reflectance patch vectors belonging to patches of size  $3 \times 3 \times 8$ acquired from a single date (March 30, 2014) as samples. On the other hand, pixel-based single-imagery NN system uses only 8-dimensional TOA reflectance center pixel vectors belonging to patches extracted for patch-based single-imagery NN system. In both patch-based and pixel-based multi-images NN system,  72-dimensional (3*3*8) flattened TOA reflectance patch vectors and 8-dimensional TOA reflectance center pixel vectors acquired from four different dates (February 10, 2014; March 14, 2014; March 30, 2014; and January 28, 2015) as samples. The proposed PB-RNN system achieves 97.21\% in the overall accuracy and 0.967 in Kappa index. For the individual categories, this new system achieves PA and UA values more
than 94\% for all classes and the mean of conditional kappa values
belonging to 8 different classes (Mean-Kappa) is 0.97 with minimum 0.94
conditional kappa index. In some classes, the system achieves significant
improvements. For example barren land, cropland, high intensity urban and woody wetland have (98.04\%, 100\%,
1.00), (98.1\%, 98.1\%, 0.98), (97.47\%, 97.47\%, 0.97) and (97.99\%, 96.53\%, 0.96) as PA, UA and conditional kappa index
values, respectively. The proposed system is able to achieve  good results for several spectrally complex classes, such as the low intensity urban and
cropland. Table~\ref{matrix_rnn} shows the complete error matrix of
system with calculated OA, KAPPA of the overall system and PA, UA,
conditional kappa  for all individual classes separately. \\

The proposed PB-RNN system achieves better results than pixel-based RNN system, pixel-based single-imagery NN system, pixel-based multi-images NN system, patch-based single-imagery NN system and patch-based multi-images NN system. Comparative results are summarized in
Table~\ref{comparison}. Figure~\ref{refer_comp} shows the comparison
to reference map from the classification results of six different
systems. The proposed system gets 97.21\% OA, 0.97 KAPPA, 0.97
Mean-Kappa; which achieves (9.56\%, 0.11, 0.13), (19.58\%, 0.23, 0.24), (21.76\%, 0.26, 0.27), (30.81\%, 0.37, 0.36) and (32.47\%, 0.39, 0.39)
improvements over pixel-based RNN system, patch-based multi-images NN system, patch-based single-imagery NN system, pixel-based multi-images NN system and pixel-based single-imagery NN system, respectively. Table~\ref{comparison} also shows that
there are significant enhancements not only in the overall but
also in all the individual categories. Table~\ref{comparison} shows
that the proposed PB-RNN system achieves a minimum 0.15 increase in conditional kappa
values for five classes with the maximum 0.19 when comparing with the pixel-based RNN system. In comparison to the
patch-based multi-images and single-imagery NN system, the proposed system achieves a minimum
0.20 and 0.22 increase in conditional kappa values for six classes with maximum 0.42 and 0.46 respectively.  In case of pixel-based multi-images and single-imagery NN system, the proposed system achieves a minimum
0.31 and 0.37 increase in conditional kappa values for all the classes except water with maximum improvements of 0.45 and 0.50 respectively. The proposed PB-RNN system
shows substantial improvements in the conditional kappa results for
hard-to-classify classes also. For example, in case of low intensity urban and
cropland, there are (0.16, 0.03), (0.27, 0.20), (0.28, 0.20), (0.40, 0.41) and (0.40, 0.41) improvements in conditional kappa values over pixel-based RNN system, patch-based multi-images NN system, patch-based single-imagery NN system, pixel-based multi-images NN system and pixel-based single-imagery NN system, respectively.\\

Several previous studies have suggested single hidden layer neural networks
perform  better for classification of remote
sensing images~\cite{kanellopoulos1997strategies,shupe2004cover}. Therefore,
we have used only one fully connected hidden layer between input and
softmax (outer) layer for both pixel and patch single-imagery NN system; the hidden
layer consists of 200 neurons. Patch-based single-imagery NN system uses 72-dimensional (3*3*8) flattened TOA reflectance patch vectors as inputs and pixel-based single-imagery NN system uses only 8-dimensional TOA reflectance center pixel vectors. In case of both multi-images NN systems, we have fused four single-imagery neural network classifiers belonging to different dates together by using joint probabilities of classes at the four dates~\cite{bruzzone1999neural}. Similar to the
proposed PB-RNN system, we have used Google's
tensorflow~\cite{tensorflow2015-whitepaper} and Quadro K5200 GPU to
implement all the other networks also.  As shown in
Table~\ref{comparison} and Figure~\ref{refer_comp}, 
both patch and pixel multi-images NN systems show only slight improvement over their respective single-imagery NN systems in  OA, KAPPA, Mean-Kappa of the overall systems and PA, UA, conditional kappa for all individual classes. Unlike the RNN systems, the multi-images NN systems consider each imagery independent to each other and is unable to utilize inherit dependency of multi-temporal remote sensing data. In patch-based multi-images NN system, there is 2.18\% improvement in OA, 0.03  improvement in KAPPA, and 0.03 improvement in Mean-Kappa, over patch-based single-imagery NN system. In pixel-based multi-images NN system, there is 1.66\% improvement in OA, 0.02  improvement in KAPPA, and 0.03 improvement in Mean-Kappa, comparing to pixel-based single-imagery NN system. Considering spatial information, patch-based single-imagery NN system improves significantly over both pixel-based single-imagery NN system and multi-images NN system. In the patch-based single-imagery NN system, there are 10.71\% and 9.05\% improvement in OA, 0.13 and 0.11 improvement in KAPPA, and 0.12 and 0.09 improvement in Mean-Kappa, comparing to pixel-based single-imagery NN system and multi-images NN system, respectively. However, without any spatial information the pixel-based RNN system is able to utilize the information of the inherit dependency of multi-temporal remotely sensed data and the invaluable spectral patterns associated with specific classes over time improves significantly over both patch-based NN systems. In pixel-based RNN system, there is 10.02\% improvement in OA, 0.12  improvement in KAPPA, and 0.11 improvement in Mean-Kappa, over patch-based multi-images NN system.  The same weighted stratified sampling is used for accuracy
assessments in all these networks too. The details of the error matrices, OA, KAPPA, PA, UA and and conditional kappa
are given in Table~\ref{matrix_rnn_pixel} for pixel-based RNN system, in Table~\ref{matrix_pixel_single} for pixel-based single-imagery NN system, in Table~\ref{matrix_pixel_multi} for pixel-based multi-images NN system, in Table~\ref{matrix_patch_single} for patch-based single-imagery NN system and in Table~\ref{matrix_patch_multi} for patch-based multi-images NN system, respectively. The substantial improvements will lead to more accurate land cover data that are essential for many applications (e.g., agriculture monitoring, energy development and resource exploration).

\begin{table*}\caption{\bf Error Matrix using the Pixel-Based Single-Imagery NN system.}\label{matrix_pixel_single}
\begin{center}
   \scalebox{0.545}{
    \begin{tabular}{  | c |  c | c |  c | c |  c |  c | c |  c | c | c | c | c |}
    \hline
     \multirow{4}{*}{Classified Data}& \multicolumn{9}{|c|}{Reference Data } &  \multicolumn{3}{|c|}{Accuracy and Conditional Kappa}\\  \cline{2-13}
      & { High} & { Low}  & { Barren} & { Forest} & {Crop} &{ Woody}& { Emergent }  &{ Water}& { Row}  & { Producer's} & { User's} & { Conditional} \\ 
     & { Intensity} & { Intensity} & { Land} & {} & {land} &{  Wetland}& {Herbaceous}   &{ } & { Total} & { Accuracy} & {Accuracy} & { Kappa} \\  
     & { Urban} & { Urban}  &{ } & {} &{} & {} & { Wetland} & {} & { } & { (PA \%)} & {(UA \%)} & { } \\  \hline
      High Intensity Urban & {\bf 130} & 36 &   4 &   8 &  11 &   4 &   4 &  13 & 210 & 76.02 & 61.90 & 0.54 \\ \hline
     Low Intensity Urban  & 8 & {\bf 29} & 0 &   0 &   6 &   1 &   2 &   4 & 50 & 30.85 & 58 & 0.54\\ \hline
     Barren Land & 5  &  0  & {\bf 27} & 8 &   5 &   1 &   1 &   3 & 50  & 69.23 & 54 & 0.52\\ \hline
    Forest  & 1  &  3  &  4 & {\bf 50} & 9  & 14  & 10  &  1 & 92  & 40 & 54.35 & 0.48\\ \hline
   Cropland & 10  & 10   & 3   & 10 & {\bf 71} & 3 &   5 &   3 & 115 & 62.28 & 61.74 & 0.57\\ \hline
   Woody Wetland  & 6  & 14  &  0  & 43  &  7 & {\bf 170} & 23  &  7 & 270 & 83.74 & 62.96 & 0.53\\ \hline
    Emergent Herbaceous   & \multirow{2}{*}4 & \multirow{2}{*}0 &  \multirow{2}{*}0 & \multirow{2}{*}6 & \multirow{2}{*}3 & \multirow{2}{*}5 & \multirow{2}{*}{\bf 30}   & \multirow{2}{*}2 & \multirow{2}{*}{50}  & \multirow{2}{*}{40} & \multirow{2}{*}{60} & \multirow{2}{*}{0.57}\\ 
      Wetland &  &  &  &  &  &  &  &  &   &  &  & \\ \hline
       Water  & 7  &  2  &  1   & 0   & 2  &  5  &  0  & {\bf 130} & 147 & 79.75 & 88.44 & 0.86 \\ \hline

    Column Total  & 171 & 94 & 39 & 125 & 114 & 203 & 75  & 163 & {\bf 984} & \multicolumn{3}{|c|}{}\\ \hline
 \multicolumn{13}{|c|}{ Overall Accuracy(OA): {\bf 64.74\%};  Overall Kappa(KAPPA): {\bf 0.583} } \\  
    \hline
     
    \end{tabular}
}
\end{center}
\end{table*}
\begin{table*}\caption{\bf Error Matrix using the Pixel-Based Multi-Images NN system.}\label{matrix_pixel_multi}
\begin{center}
   \scalebox{0.545}{
    \begin{tabular}{  | c |  c | c |  c | c |  c |  c | c |  c | c | c | c | c |}
    \hline
     \multirow{4}{*}{Classified Data}& \multicolumn{9}{|c|}{Reference Data } &  \multicolumn{3}{|c|}{Accuracy and Conditional Kappa}\\  \cline{2-13}
      & { High} & { Low}  & { Barren} & { Forest} & {Crop} &{ Woody}& { Emergent }  &{ Water}& { Row}  & { Producer's} & { User's} & { Conditional} \\ 
     & { Intensity} & { Intensity} & { Land} & {} & {land} &{  Wetland}& {Herbaceous}   &{ } & { Total} & { Accuracy} & {Accuracy} & { Kappa} \\  
     & { Urban} & { Urban}  &{ } & {} &{} & {} & { Wetland} & {} & { } & { (PA \%)} & {(UA \%)} & { } \\  \hline
      High Intensity Urban & {\bf 136} & 30  & 6 & 7  & 9 & 2 & 5 & 17 & 212 & 79.53 & 64.15 & 0.57 \\ \hline
     Low Intensity Urban  & 7 & {\bf 29} & 1 & 2 & 6 & 1 &  1  & 3 & 50 & 36.71 & 58 & 0.54\\ \hline
     Barren Land & 3 & 1  & {\bf 34} & 5 & 5 & 0 & 0  & 2 & 50  & 66.67 & 68 & 0.66\\ \hline
    Forest  & 3 & 1 & 2 & {\bf 46} & 7 & 12 & 7  & 0 & 78  & 35.66 & 58.97 & 0.53\\ \hline
   Cropland & 9 & 6 & 5 & 15 & {\bf 75} & 8 & 2 & 0 & 120 & 63.03 & 62.5 & 0.57\\ \hline
   Woody Wetland  & 6 & 10 & 0 & 46 & 12 & {\bf 173} & 33  & 5 & 285 & 85.22 & 60.7 & 0.51\\ \hline
    Emergent Herbaceous   & \multirow{2}{*}1 & \multirow{2}{*}0 &  \multirow{2}{*}1 & \multirow{2}{*}7 & \multirow{2}{*}3 & \multirow{2}{*}5 & \multirow{2}{*}{\bf 33}   & \multirow{2}{*}0 & \multirow{2}{*}{50}  & \multirow{2}{*}{40.74} & \multirow{2}{*}{66} & \multirow{2}{*}{0.63}\\ 
      Wetland &  &  &  &  &  &  &  &  &   &  &  & \\ \hline
       Water  & 6 & 2 & 2 & 1 & 2 & 2 & 0 & {\bf 134} & 149 & 83.23 & 89.93 & 0.88 \\ \hline

    Column Total   & 171 & 79 & 51 & 129 & 119 & 203 & 81  & 161 & {\bf 994}  & \multicolumn{3}{|c|}{}\\ \hline
 \multicolumn{13}{|c|}{ Overall Accuracy(OA): {\bf 66.40\%};  Overall Kappa(KAPPA): {\bf 0.602} } \\  
    \hline
     
    \end{tabular}
}
\end{center}
\end{table*}

\begin{table*}\caption{\bf Error Matrix using the Patch-Based Single-Imagery NN system.}\label{matrix_patch_single}
\begin{center}
   \scalebox{0.545}{
    \begin{tabular}{  | c |  c | c |  c | c |  c |  c | c |  c | c | c | c | c |}
    \hline
     \multirow{4}{*}{Classified Data}& \multicolumn{9}{|c|}{Reference Data } &  \multicolumn{3}{|c|}{Accuracy and Conditional Kappa}\\  \cline{2-13}
      & { High} & { Low}  & { Barren} & { Forest} & {Crop} &{ Woody}& { Emergent }  &{ Water}& { Row}  & { Producer's} & { User's} & { Conditional} \\ 
     & { Intensity} & { Intensity} & { Land} & {} & {land} &{  Wetland}& {Herbaceous}   &{ } & { Total} & { Accuracy} & {Accuracy} & { Kappa} \\  
     & { Urban} & { Urban}  &{ } & {} &{} & {} & { Wetland} & {} & { } & { (PA \%)} & {(UA \%)} & { } \\  \hline
      High Intensity Urban & {\bf 135} & 28  & 3 & 5 & 5 & 3 & 0 & 3 & 182 & 80.36 & 74.18 & 0.69 \\ \hline
     Low Intensity Urban  & 12 & {\bf 48} & 1 & 2 & 5 & 0 & 0   & 1 & 69 & 51.06 & 69.57 & 0.66\\ \hline
     Barren Land & 4 & 0  & {\bf 31} & 4 & 4 & 1 & 2  & 4 & 50  & 70.45 & 62 & 0.60\\ \hline
    Forest  & 3 & 9 & 2 & {\bf 72} & 10 & 15 & 14  & 0 & 125  & 63.72 & 57.6 & 0.52\\ \hline
   Cropland & 8 & 3 & 4 & 1 & {\bf 86} & 2 & 2 & 1 & 107 & 74.14 & 80.37 & 0.78\\ \hline
   Woody Wetland  & 4 & 6 & 0 & 27 & 2 & {\bf 168} & 13  & 2 & 222 & 84.85 & 75.68 & 0.69\\ \hline
    Emergent Herbaceous   & \multirow{2}{*}1 & \multirow{2}{*}0 &  \multirow{2}{*}0 & \multirow{2}{*}1 & \multirow{2}{*}2 & \multirow{2}{*}8 & \multirow{2}{*}{\bf 37}   & \multirow{2}{*}1 & \multirow{2}{*}{50}  & \multirow{2}{*}{54.41} & \multirow{2}{*}{74} & \multirow{2}{*}{0.72}\\ 
      Wetland &  &  &  &  &  &  &  &  &   &  &  & \\ \hline
       Water  & 1 & 0 & 3 & 1 & 2 & 1 & 0 & {\bf 152} & 160 & 92.68 & 95 & 0.94 \\ \hline

    Column Total  & 168 & 94 & 44 & 113 & 116 & 198 & 68  & 164 & {\bf 965} & \multicolumn{3}{|c|}{}\\ \hline
 \multicolumn{13}{|c|}{ Overall Accuracy(OA): {\bf 75.54\%};  Overall Kappa(KAPPA): {\bf 0.712} } \\  
    \hline
     
    \end{tabular}
}
\end{center}
\end{table*}
\begin{table*}\caption{\bf Error Matrix using the Patch-Based Multi-Images NN system.}\label{matrix_patch_multi}
\begin{center}
   \scalebox{0.545}{
    \begin{tabular}{  | c |  c | c |  c | c |  c |  c | c |  c | c | c | c | c |}
    \hline
     \multirow{4}{*}{Classified Data}& \multicolumn{9}{|c|}{Reference Data } &  \multicolumn{3}{|c|}{Accuracy and Conditional Kappa}\\  \cline{2-13}
      & { High} & { Low}  & { Barren} & { Forest} & {Crop} &{ Woody}& { Emergent }  &{ Water}& { Row}  & { Producer's} & { User's} & { Conditional} \\ 
     & { Intensity} & { Intensity} & { Land} & {} & {land} &{  Wetland}& {Herbaceous}   &{ } & { Total} & { Accuracy} & {Accuracy} & { Kappa} \\  
     & { Urban} & { Urban}  &{ } & {} &{} & {} & { Wetland} & {} & { } & { (PA \%)} & {(UA \%)} & { } \\  \hline
          High Intensity Urban & {\bf 141} & 21  & 4 & 9 & 4 & 2 & 0 & 5 & 186 & 84.94 & 75.81 & 0.71 \\ \hline
     Low Intensity Urban  & 10 & {\bf 47} & 2 & 2 & 4 & 2 & 1   & 0 & 68 & 64.38 & 69.12 & 0.67\\ \hline
     Barren Land & 3 & 0  & {\bf 33} & 4 & 4 & 0 & 2  & 4 & 50  & 71.74 & 66 & 0.64\\ \hline
    Forest  & 1 & 2 & 4 & {\bf 78} & 8 & 21 & 12  & 0 & 126  & 63.41 & 61.9 & 0.56\\ \hline
   Cropland & 2 & 1 & 1 & 5 & {\bf 89} & 5 & 7 & 1 & 111 & 79.46 & 80.18 & 0.78\\ \hline
   Woody Wetland  & 4 & 1 & 0 & 21 & 1 & {\bf 170} & 19  & 3 & 219 & 84.58 & 77.63 & 0.72\\ \hline
    Emergent Herbaceous   & \multirow{2}{*}0 & \multirow{2}{*}0 &  \multirow{2}{*}1 & \multirow{2}{*}4 & \multirow{2}{*}2 & \multirow{2}{*}1 & \multirow{2}{*}{\bf 42}   & \multirow{2}{*}0 & \multirow{2}{*}{50}  & \multirow{2}{*}{50.6} & \multirow{2}{*}{84} & \multirow{2}{*}{0.83}\\ 
      Wetland &  &  &  &  &  &  &  &  &   &  &  & \\ \hline
       Water  & 5 & 1 & 1 & 0 & 0 & 0 & 0 & {\bf 153} & 160 & 92.17 & 95.62 & 0.95 \\ \hline

    Column Total  & 166 & 73 & 46 & 123 & 112 & 201 & 83  & 166 & {\bf 970} & \multicolumn{3}{|c|}{}\\ \hline
 \multicolumn{13}{|c|}{ Overall Accuracy(OA): {\bf 77.63\%};  Overall Kappa(KAPPA): {\bf 0.737} } \\  
    \hline
     
    \end{tabular}
}
\end{center}
\end{table*}
\section{ Conclusion and future work} \label{sec5}

In this paper, we have proposed a new patch-based RNN system tailored for land cover classification. The proposed system uses new features to exploit the complete multi-temporal, multi-spectral and spatial information together for land cover mapping. Specifically
for Landsat data, we have computed multi-temporal-spectral-spatial samples representing sequence of 23 patches of size $3 \times 3 \times 8$ belonging to the center pixel location of a distinct $3 \times 3$ window over the whole year from multi-temporal-spectral remote sensing imagery. The proposed system is capable of utilizing the spatial information of the inherit dependency of multi-temporal remotely sensed data and the invaluable spectral patterns associated with specific classes over time; by using input gate in the LSTM cell, we are able to deal cloud/shadow pixels by restricting these pixel vectors to let through the input gate. The proposed system is compared to pixel-based RNN system, pixel-based single-imagery NN system, pixel-based multi-images NN system, patch-based single-imagery NN system and patch-based multi-images NN system.. The classification results show that the proposed system achieves significant improvements in both the overall and categorical classification accuracies.\\

There are further changes that could lead to further improvements. For example, classification accuracy could improve further by creating hierarchical structure classification on the top of the proposed system using different sizes of patches for the same center
pixel. Convolution neural network (CNN) can be incorporated with this PB-RNN system to improve the performance even more. We believe that we can develop season-based classifier using the same technique. These and other parameter choices are being
investigated further.

 %we need an
%automatic system to extract training samples belonging to individual
%categories separately.

\clearpage
\bibliographystyle{plain}
\bibliography{myreference}
\end{document}